\documentclass{article}

\usepackage[numbers]{natbib}

\usepackage {arxiv}
\usepackage{cite}
\usepackage{floatrow}
\usepackage{amsmath,amssymb,amsfonts}
\usepackage{caption}
\usepackage{algorithm}
\usepackage{algorithmic}
\usepackage{textcomp}
\usepackage[table]{xcolor}
\usepackage{float}
\usepackage{booktabs}
\usepackage{enumitem}
\usepackage{setspace}
\usepackage[]{graphicx}
\usepackage{amsmath}
\usepackage{amssymb}
\usepackage{epsfig}
\usepackage{epstopdf}

\usepackage{subfigure}
\usepackage{caption}
\usepackage{comment}
\usepackage{multirow}

\title{\LARGE \bf
Missing Value Estimation using Clustering and Deep Learning within Multiple Imputation Framework}

\author{Manar D. Samad and Sakib Abrar\\
Department of Computer Science\\
 Tennessee State University\\
 Nashville, TN, USA\\
\texttt{msamad@tnstate.edu} \\
\and
 \bf {Norou Diawara}\\
 Department of Mathematics and Statistics\\
  Old Dominion University\\
   Norfolk, VA, USA }
   
\begin{document}



\maketitle

\begin{abstract}
Missing values in tabular data restrict the use and performance of machine learning, requiring the imputation of missing values. The most popular imputation algorithm is arguably multiple imputations using chains of equations (MICE), which estimates missing values from linear conditioning on observed values. This paper proposes methods to improve both the imputation accuracy of MICE and the classification accuracy of imputed data by replacing MICE’s linear conditioning with ensemble learning and deep neural networks (DNN). The imputation accuracy is further improved by characterizing individual samples with cluster labels (CISCL) obtained from the training data. Our extensive analyses involving six tabular data sets, up to 80\% missingness, and three missingness types (missing completely at random, missing at random, missing not at random) reveal that ensemble or deep learning within MICE is superior to the baseline MICE (b-MICE), both of which are consistently outperformed by CISCL. Results show that CISCL + b-MICE outperforms b-MICE for all percentages and types of missingness. Our proposed DNN based MICE and gradient boosting MICE plus CISCL (GB-MICE-CISCL) outperform seven other baseline imputation algorithms in most experimental cases. The classification accuracy on the data imputed by GB-MICE is improved by proposed GB-MICE-CISCL imputed data across all missingness percentages.  Results also reveal a shortcoming of the MICE framework at high missingness ($>$50\%) and when the missing type is not random. This paper provides a generalized approach to identifying the best imputation model for a data set with a missingness percentage and type. 
\end{abstract}

\section{INTRODUCTION}
The inevitable presence of missing values is one major challenge in building machine learning-based predictive models using real-world data. Missing values in data are often imputed by an arbitrary imputation strategy to enable machine learning-based predictive modeling.  However, imputation algorithms generate new synthetic values, which may alter the data distribution and mislead or inflate data-driven predictive models~\citep{leke2015proposition}. Therefore, the quality, statistics, and proportion of imputed values play a critical role in subsequent training and outcomes of predictive models. Many missing value imputation (MVI) studies limit their experiments on iteratively imputing a training data set without evaluating such models on imputing separate test data or classification tasks~\citep{CEVALLOSVALDIVIEZO2015163}. The type and pattern of data missingness are rarely analyzed to determine a suitable imputation model. On the other hand, the machine learning literature evaluates classification algorithms using benchmark tabular data sets that are clean, curated, complete, and free from the challenge of data missingness. Hence, there is a gap between the study of MVI and the machine learning literature innovating new data-driven predictive models. The problem of MVI primarily draws interests in statistics and domain research (bioinformatics, health science, sensors), aiming to bring machine learning-based solutions to respective domain data with missing values. A substantial portion of the MVI literature proposes imputation models for a single domain data set considering a limited percentage of missingness.   

In recent years, machine learning-based solutions to MVI problems have emerged, promising to outperform traditional statistical and matrix completion approaches. Existing data imputation strategies can be broadly categorized into single imputation (mean, median), non-MICE imputation (matrix factorization, k-nearest neighborhood), multiple imputations (using chained equations), imputation using ensemble learning (random forest), and deep learning (generative models, autoencoders). However, little or no study extensively compares all these model categories with different domain data, missingness types, and percentages, as we demonstrate in this paper. Instead, there is a consensus in the literature that multiple imputations using chained equations (MICE) are the most accurate models for data imputation because of its conditional modeling of missing variables on other variables~\citep{VanBuuren}. Therefore, we take the conventional MICE model as the baseline to compare the performance of advanced machine learning models embedded in the MICE framework, including ensemble learning, deep learning, and our proposed model. We hypothesize that a hybrid framework that trains ensemble or deep learning models within the multiple imputation framework will improve the MVI accuracy. Additionally, we propose a novel approach to improve these hybrid model performance by data clustering and infusing the cluster information into individual samples for minimizing the variability in MVI estimates.   

The remainder of the article is organized as follows. Section II provides an overview of MVI approaches and a literature review of the existing MVI algorithms. Section III discusses the proposed algorithm and experimental steps. Section IV provides the findings following the proposed experimentation. Section V discusses the findings with explanations and limitations of our work. The article concludes in Section VI with some future work.  
\begin{table}[t]
\caption{Recent articles on missing value imputation. Three types: missing completely at random (MCAR), missing at random (MAR), and missing not at random (MNAR).}
\begin{tabular}{@{}llcl@{}}
\toprule
                        & Missing & 
Max \%  
& Data       \\

&type&of missing&domain  \\
\midrule
Nikfalazar et al.~\citep{Nikfalazar2020}  & MCAR & 10\%       & Multiple\\
Awan et al.~\citep{Awan_Elsevier_2021}& MCAR & 20\% & Multiple\\
Camino et al.~\citep{Camino2019}& MCAR   & 80\%  &
Multiple \\
Wang et al.~\citep{Wang2021} & Three types & 30\% & Multiple \\
Chen et al.~\citep{chen2020comparison}
& MNAR, MAR& 40\%       & Single \\
Ward et al.~\citep{ward2020approaches}  & MNAR, MAR& 27\% &  Single \\
Jones et al.~\citep{beaulieu2018characterizing} & Three types & 50\% & Single \\
\bottomrule
\end{tabular}
\vspace{-15pt}
\label{literature}
\end{table}

\section {Background review}

This section reviews the theoretical underpinning of the MICE framework and the state-of-the-art literature in MVI studies.  

\subsection {Missing values in tabular data} Let's denote a data matrix, X = \{$X_{obs}$, $X_{miss}$\} where $X_{obs}\subset X$ includes the observed or existing values, and $X_{mis}\subset X$ represents the missing values. A response indicator matrix (R = [$r_{ij}$]) identifies the location of missing values in X, taking $r_{ij}$ = 0 for missing entries and $r_{ij}$ = 1 for observed. A conditional probability distribution of R = 0 (missingness) given X can be defined as $Pr (R = 0 | X_{obs} , X_{mis})$, which is used to model three types of missingness: missing completely at random (MCAR), missing at random (MAR), and missing not at random (MNAR). First, MCAR occurs when the probability of missingness is completely independent of the observed ($X_{obs}$) or missing ($X_{miss}$) data as:
\begin{equation}
    Pr (R =0 ) = Pr (R=0| X_{obs} , X_{mis}).
\end{equation}
Second, when the probability of missingness depends on the observed data, it represents MAR as:
\begin{equation}
    Pr (R =0| X_{obs}) = Pr (R =0| X_{obs} , X_{mis}).
\end{equation}
Third, MNAR is identified when the missing probability depends on the unobserved or missing values themselves as follows. 
\begin{equation}
    Pr (R=0| X_{mis}) = Pr (R=0| X_{obs} , X_{mis}).
\end{equation}

The literature has very limited studies that comprehensively analyze MVI algorithms for all missingness types, higher percentages of missingness, and data from multiple domains (Table~\ref{literature}). Most studies have considered only the MCAR type because of the convenience of rendering completely random missing positions~\citep{Nikfalazar2020, Gonzalez-Vidal2020, HEGDE2019100275}. In a survey of the MVI literature, only 12 out of 111 articles (10.8\%) have studied imputations of data with more than 50\% missing values~\citep{Lin2020}. For example, Nikfalazar et al. have tested MCAR type missingness for up to 10\% of missing data~\citep{Nikfalazar2020}. In practice, MAR and MNAR types are more common than the MCAR type. Table~\ref{literature} highlights missingness types, percentages, and data domains of recent MVI studies. Intuitively, an MVI algorithm will be lesser accurate on data with a higher percentage of missing values and vice versa. Therefore, the robustness of MVI algorithms can be evaluated over a range of missing value percentages. For example, Jones et al. have imputed a large electronic health records (EHR) data set for up to 50\% missingness~\citep{beaulieu2018characterizing}.  Our previous study on a large EHR data set (sample size $>$ 300,000) reveals that the most important predictor of patient mortality (out of 149 variables) can be missing for more than 50\% of the samples~\citep{Samad2018}. Similar predictor variables with over 50\% of missing values are often excluded before imputation and predictive modeling, limiting our ability to yield important insights from data. Therefore, the trade-off between data imputation and exclusion at a high missing percentage is critical for optimizing machine learning model performance.

\subsection {Statistical imputation of missing values.} The field of MVI is predominantly researched by statisticians~\citep{VanBuuren}. There are varying statistical patterns in missing values (e.g., monotone, k-monotone, arbitrary)~\citep{VanStein2016} that may benefit from targeted imputation algorithms. One of the most successful MVI algorithms is built on the statistical concept of multiple imputations (MI)~\citep{VanBuuren, Nassiri2020}, such as multiple imputations using chained equations (MICE)~\citep{Resche-Rigon2018, Luo2018}. As opposed to single value imputations (mean, median, or other point estimates such as k-nearest neighbors), multiple imputations generate multiple samples of imputed data sets. The variability across multiple imputed data sets captures the statistical uncertainty in data imputations, improving the imputation accuracy. The MICE algorithm estimates missing values of individual variables using observed values of other variables, which intuitively models the MAR type.
\begin{algorithm}[t]
\caption{Multiple imputation with chained equations}
\begin{algorithmic}
\STATE Input: Data matrix $X = \{X_{obs}, X_{miss}\}$
\STATE Output: Imputed data matrix, $X = \{X_{obs}, X_{imputed}\}$
\FOR {i = 1 $\rightarrow$ 5 (multiple imputations)} 
\STATE \{$\bar{X}, \sigma^2\} \leftarrow {X_{obs}}$
\STATE Initial imputation, $X^0_{miss} \leftarrow$ $\mathcal{N}$~($\bar{X},\sigma^2)$
\FOR {t = 1 $\rightarrow$ N} 
\STATE Estimate $W^t$, Equation 6
\STATE Estimate  $X_{miss}^t$, Equation 7 
\STATE $X_{miss}^t \sim P(X_{miss}^t|X_{obs},X_{miss}^{-t})$
\ENDFOR
\STATE $X_{imputed}^i$ $\leftarrow$ $X_{miss}^{N}$
\ENDFOR
\RETURN $X_{imputed}$ $\leftarrow$ Aggregate ($X_{imputed}^i$)
\end{algorithmic}
\end{algorithm}
Let a multivariate data matrix X = [$x_{ij}$] = \{ $x_1, x_2,...,x_n$ \}$\in \Re^{n \times d}$ with missing values has $n$ samples and $d$ variables. In MICE, the j-th variable $x_j \in \Re^n$ can be regressed from observed and imputed values of other $d-1$ variables $x_k$ as below.
\begin{equation}
     x_j = f (x_k, W),~k \neq j, 
\end{equation}
where j-th ($x_j$) variable is regressed using a vector of parameters $W$ as below. 
\begin{equation}
W =[ w_1, w_2, ..., w_{d-1} ]^T.  
\end{equation}
The regression parameter vector $W\in\Re^{d-1}$ is drawn by a Gibbs sampler as shown in Equation 6. Given an estimation of $\hat{W_j}$ for variable $x_j$, missing values of this variable can be sampled from the conditional distribution in Equation 7. 
\begin{eqnarray}
{ \hat{W_j^t}} \sim Pr (W^t_j \vert x_j^{obs},x_{(k<j)}^t, x^{t-1}_{(k>j)}),\\
x_j^t \sim Pr (x_j^{miss} \vert x_j^{obs},x_{(k<j)}^t, x^{t-1}_{(k>j)},\hat{W_j^t}). \end{eqnarray}
Here, $t$ is the iteration index. Both 
$\hat{W_j^t}$ and $x_j^t$ are iteratively updated using a chain of $d$ equations (for $d$ variables) until convergence following Algorithm 1.  These sampling distributions generate multiple imputed data sets for multiple imputations. The conditional sampling distributions (Equations 6 and 7) of the MICE framework can be explained by the Bayesian approach. In the Bayesian approach, the posterior distribution of imputation model parameter (W) given observed data ($X_{obs}$) and estimated missing values ($y_{miss}$) is likelihood $P(y_{miss} \vert X_{obs}, W)$ times the prior distribution, P(W). 
\begin{equation}
P(W \vert y_{miss}, X_{obs}) = P(y_{miss} \vert X_{obs}, W) P(W) 
\end{equation}
We set the prior distribution of W as Gaussian. 
\begin{equation}
P(W) = N(\mu_0, \sigma^2_0)
\end{equation}
Considering zero mean $\mu_0$=0 Gaussian distribution and ignoring the normalizing constant, the prior becomes:
\begin{align*}
P(W) &= \frac{1}{\sqrt{2 \sigma^2_0 \pi}} \, \exp \left( -\frac{(W - \mu_0)^2}{2 \sigma^2_0} \right) \\[10pt]
                             &\propto \exp \left( -\frac{W^2}{2 \sigma^2_0} \right)
\end{align*}
Therefore, the posterior (Equation 8) becomes as follows. 
\begin{eqnarray}
P(W \vert y_{miss}, X_{obs}) \nonumber\\
                      \propto \exp \left( -\frac{(y_{miss} - W^T X_{obs})^2}{2 \sigma^2} \right) \, \exp \left( -\frac{W^2}{2 \sigma^2_0} \right)
\end{eqnarray}
Taking log posterior we obtain:
\begin{eqnarray}
\log P(W \vert y_{miss}, X_{obs}) \nonumber\\ 
\propto -\frac{1}{2 \sigma^2}(y_{miss} - W^T X_{obs})^2 - \frac{1}{2 \sigma^2_0} W^2 \nonumber \\[5pt]
                           = -\frac{1}{2 \sigma^2} \Vert y_{miss} - W^TX_{obs}\Vert^2_2 - \frac{1}{2 \sigma^2_0} \Vert W \Vert^2_2
\end{eqnarray}

We assume that $\sigma^2$=1 and $\lambda=\frac{1}{\sigma_0^2}$, then our log posterior becomes:
\begin{equation}
\log P(W \vert y_{miss}, X_{obs}) \propto -\frac{1}{2} \Vert y_{miss} - W^TX_{obs}\Vert^2_2 - \frac{\lambda}{2} \Vert W \Vert^2_2
\end{equation}
That is, the log posterior of the imputation model is equivalent to the objective function for ridge regression. However, this result yields a point estimate instead of a distribution. For the full Bayesian approach during testing, we 
marginalize the posterior predictive distribution. Given trained parameters (W) and a new data point with missing values ($x'$), the posterior distribution of imputed values given ($x'$) and trained imputation model (W)  is $P(y' \vert x', W)$. This distribution is marginalized by the posterior of weight distribution $P(W \vert y_{miss}, X_{obs})$ as below.
\begin{eqnarray}
P(y' \vert y_{miss}, X_{obs}) &=& \int_W P(y' \vert x', W) P(W \vert y_{miss}, X_{obs}) \nonumber \\[5pt]
                       &=& \mathbb{E}_W \left[ P(y' \vert x', W) \right]
\end{eqnarray}
In practice, each component of $W$ is a real number, which makes the problem of enumerating all possible values of $W$ intractable. Therefore, approximation methods such as Monte Carlo Markov chains (MCMC) are used, which are computationally more expensive than point estimates but more accurate.
\begin{table*}[t]
\caption{A summary of the six data sets used to evaluate the performance of missing value imputation algorithms.}
\begin{tabular}{lccccc}
 \toprule
Data set                             & Sample & \# of & \# of target & Type  & Range\\  
&size&variables&classes&\\
 \midrule
Breast cancer  & 569     & 30       & 2            & Diagnostic      & 0 to 4254 \\
Dermatology                         & 358     & 35       & 6            & Histopathological & 0 to 9 \\
SkillCraft1 master table          & 3395    & 19       & NA           & Game  & 0 to 1000000                       \\
Wine quality 
& 4898    & 11       & 7            & Food    & 0 to 440                    \\
Default of credit card clients 
& 30000   & 21\footnote{Three categorical variables (sex, education, marriage) are excluded.}       & 2            & Financial     & -339603 to 1684259              \\
Mice protein expression 
& 552     & 77       & 8            & Proteomic     & 0 to 8.5              \\
 \bottomrule
\end{tabular}
\vspace{-10pt}
\label{tab01}
\end{table*}
\begin{figure}[t]
\vspace{0pt}
	\includegraphics [width=4cm]{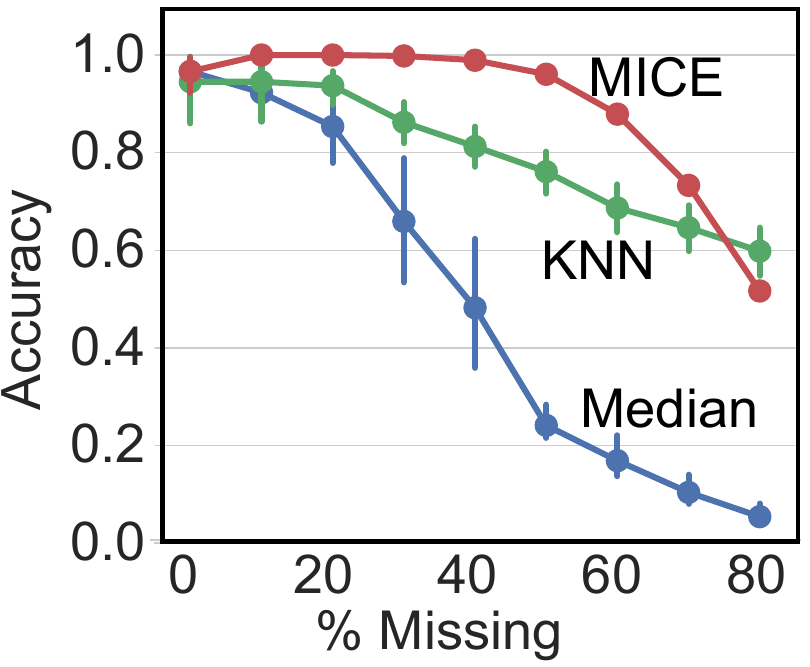}
	\vspace{-5pt}
	\caption{MICE imputed data have shown to yield superior and robust 
	classification accuracy compared to single imputation methods, e.g., median and K-nearest neighbor.}
	\label{figMICE}
\end{figure}
The imputed missing values using linear regression models in multiple imputation framework (Algorithm 1) yield better classification accuracy than those imputed by single imputations such as K-nearest neighbors (KNN)~(Fig.~\ref{figMICE}). The MICE algorithm is touted as one of the most successful imputation methods in state-of-the-art data-driven studies~\citep{Kose2020, Samad2018, VanBuuren}. Therefore, the MICE model has been evaluated and compared in most recent MVI studies, including~\citep{neuroISP, Nyugen_Elsevier_2021, Awan_Elsevier_2021}.
\subsection {Machine learning in data imputation}
The recent literature extends statistical MVI models with more advanced machine and deep learning (MnD) algorithms. Deep learning is proposed in the imputation of drug discovery data~\citep{Irwin2020} and non-numerical tabular data~\citep{Biessmann2018}. Recently, deep generative models ~\citep{Zhang2018, Camino2019}, convolutional neural networks~\citep{Zhuang2019}, recurrent neural networks~\citep{Sangeetha2020} have been proposed to impute data with spatial and temporal dimensions. In another study, autoencoders are trained (encoded) using complete data set to estimate (decode) missing values from their high-dimensional latent spaces~\citep{Choudhury2019}. Apart from these variants of deep models, fully connected deep neural networks with three or more hidden layers have not been studied for MVI. On the other hand, ensemble learning, such as extreme gradient boosting trees have been trained as regressors to estimate missing values for data imputation~\citep{Madhu2019}. However, these MnD-based imputation methods do not provide the benefits of multiple imputations. In line with this notion, the multiple imputation performance of the MICE algorithm is improved by replacing the standard GLM with regression trees in categorical data imputation,~\citep{Akande2017}. A random forest classifier is used in the MICE framework to impute categorical and binary variables of a traffic crash report data set~\citep{Li_elsevier_2020}. Decision tree-based regressors have been recently proposed within the MICE framework to impute missingness in brain function and anatomy data~\citep{Slade2020}. Hallam et al. have used gradient boosting (GB) tree regressors within MICE to impute geological data set with continuous variables~\citep{Hallam_elsevier_2021}. 
The existing work reports findings on particular data types (binary or categorical), target domains, and within a limited range of missing values and types. These selective approaches may not generalize well across many other variants of data and missingness. A recent survey of the MVI literature has highlighted the importance of a) choice of data sets, b) missing rates, c) missing types, and d) evaluation metrics for missing value imputation methods~\citep{Lin2020}. We hypothesize that a hybrid framework of multiple imputations and ensemble learning can improve the performance and robustness of data imputation across various data sets, missing types, and percentages compared to the existing imputation models. 
\subsection{Data imputation algorithms}
This section provides an overview of the state-of-the-art MVI algorithms. 
\subsubsection {Machine learning in MICE}
Multiple imputations by chained equations  (MICE) consist of variable-specific linear regression models, which are first seeded with initial imputation values. The chains of these equations are then iteratively updated to refine estimates of missing values (Algorithm 1). Machine learning-based non-linear regressors can replace these linear regression equations to improve the accuracy of missing value estimations~\citep{SamadMICE}. The MI part of MICE models is known to capture the variability between models, whereas ensemble learning models the variability within a model~\citep{CEVALLOSVALDIVIEZO2015163}. An example of the MICE algorithm using an ensemble of decision trees for data imputation is MICEforest \citep{stekhoven2012missforest}. The iterative update in MICEforest is performed using predictive mean matching (PMM). In PMM, the model estimated missing value is compared with other observed values in the data set to determine the neighboring samples. The observed value of the nearest neighbor is then used to impute the missing value. Therefore, PMM performs single value imputation using the observed value of the nearest neighbor instead of performing multiple imputations. Our experiment with PMM shows no improvement in imputation accuracy compared to the MI-based imputation strategy. 
\subsection {Imputation for predictive modeling}
One major objective of MVI is to retain maximum samples for predictive modeling (classification or regression) without excluding data due to missingness. The existing MVI approaches have two shortcomings in predictive modeling. First, Valdiviezo et al.~\citep{CEVALLOSVALDIVIEZO2015163} have argued that the MVI literature almost always evaluates MVI algorithms by comparing actual and imputed values. Unfortunately, many real-world data sets with missing values do not contain the ground truth actual values. Furthermore, it remains unclear how data quantity and quality, augmented by imputation algorithms, improve the performance of classification or regression models. Second, there is no established benchmark datasets for evaluating imputation algorithms. The imputation algorithms are iteratively executed and then evaluated on the same data set with missing values in a self-supervised manner. Ideally, a trained imputation model should be used to impute a standalone test data set with missing values to report the imputation accuracy. Third, a MICE model builds chains of regression equations using all heterogeneous data points at a time. We argue that the presence of heterogeneous and outlier samples may result in higher variance or errors in the final estimates of missing values. Therefore, data imputation modeling within individual clusters of homogeneous samples may reduce the variability and errors in the estimates of missing values. 
\subsection {Contributions}
Based on our literature review, the contributions of this article is as follows. First, we provide algorithms to simulate three missing value types (MAR, MCAR, and MNAR) in tabular datasets. Second, we perform extensive analyses to demonstrate the effect of missing value percentages up to 80\% across three missing value types on imputation algorithms. We hypothesize that no single imputation algorithm can be the best choice for imputing all types of missingness or data of any domain. Third, the imputation performance is reported on separate test data with missing values, and the quality of imputed data is subsequently evaluated in a supervised classification. We investigate if better imputation accuracy also translates to better classification performance. Fourth, we propose novel imputation algorithms by 1) deep learning in the MI framework and 2) ensemble learning within MI after infusing cluster information for homogeneous imputation. The information about data clusters is expected to improve errors in missing value estimation. Fifth, the performance of our proposed imputation algorithm is compared against non-MICE imputation algorithms, baseline MICE, and MICE with ensemble learning across MCAR, MAR, and MNAR types from 5\% to 80\% missing data.

\section {Methodology}

\subsection {Data sets}

We conduct all experiments on six tabular data sets obtained from the UCI machine learning data repository~\citep{Dua:2019} as shown in Table~\ref{tab01}. The datasets cover a good range of sample sizes (300-30,000), data dimensionality (11-77), and domains (diagnosis to game) to understand their effects on missing value estimation. The MVI models are trained using 70\% of the data set, and the remaining 30\% is used to evaluate the performance of trained MVI models. We standardize the data using the mean and standard deviation of individual variables in the training data set. The MVI model estimation of missing values is then transformed back to the original scale for comparing against the actual values. 
\subsection {Simulation of missingness types:} The three types of missingness (Section II-A) are simulated following the computational framework proposed in Algorithm 2. For MCAR, we randomly generate indices to remove values from the data matrix based on the missing rate. For MNAR, the random selection is made below the lower quantile and above the upper quantile of individual features. Here, the missingness is introduced to high or low values of the same features subjected to missingness. For MAR, the quantiles are determined on a separate set of observed feature values to remove corresponding values in other features. The total number of removed values is set by the number of samples times the number of features times the missing rate.          
\begin{algorithm} [t]
\vspace{5pt}
\caption{Simulation of three types of missingness.}
\begin{algorithmic}
\STATE Input: $X_{complete}$,~missing\_rate,~missing\_type
\STATE Output: $X\_{missing}$
\IF {missing\_type == MCAR}
\STATE [p, q] $\leftarrow$ random\_indices~(missing\_rate)
\STATE $X_{missing}$ $\leftarrow$ remove~($X_{complete}$, [p, q])
\RETURN $X_{missing}$
\ENDIF
\STATE low = missing\_rate / 2 
\STATE high = 100 - missing\_rate / 2 
\FOR {i = 1 $\rightarrow$ n\_features}
\STATE feature $\leftarrow$~$X_{complete} (:,i)$
\IF {missing\_type == MNAR}
\STATE low\_val (i) $\leftarrow$ Quantile~(feature, low)
\STATE high\_val (i) $\leftarrow$ Quantile~(feature, high)
\STATE missing$\leftarrow$select~(feature,~$<$low\_val $\cup$~$>$high\_val) 
\STATE $X_{missing}$ $\leftarrow$ remove~(feature, missing)
\ELSIF {missing\_type == MAR}
\STATE obs\_features$_{j \neq i} \subset$ \{1, 2, ..., n\_features\}
\FOR {j in obs\_features}
\STATE low\_val~(j) $\leftarrow$ Quantile~($X_{complete} (:,j)$,~low)
\STATE high\_val~(j) $\leftarrow$ Quantile~($X_{complete} (:,j)$,~high)
\STATE [p, q] $\leftarrow$ indices~($<$low\_val(j) $\cup$ $>$high\_val(j))
\ENDFOR
\STATE $X_{missing}$ $\leftarrow$~remove(feature, [p, q])
\ENDIF
\ENDFOR
\RETURN $X_{missing}$
\end{algorithmic}
\end{algorithm}
\subsection {Multiple imputation using chained regressors}
We exclude the target classification column from the data set before simulating the missingness and evaluating imputation algorithms. The procedure of multiple imputations (MI) is initialized by random sampling of multiple imputed datasets, which are slightly different from each other to capture uncertainties in data imputation (Section II-B). These imputed data sets are independently and iteratively updated using a chain of regressors, as shown in Algorithm 1. Buuren recommends 10 to 20 iterations to update the chains of regressors toward the final imputed values~\citep{VanBuuren}. The final imputed data set is aggregated (averaged) from multiple imputed data sets. We implement and evaluate MI frameworks developed using the baseline linear regression (LR-MICE), ensemble learning regressors (random forest, gradient boosting), and chains of deep regressors (deep neural networks). It is worth noting that our LR-MICE is the popular MICE algorithm used across many domains, including in our prior work~\citep{Samad2018}.
\subsection{Regression using ensemble learning}
Two of the most successful ensemble learning algorithms, random forest and gradient boosting trees, are used to build the regression chains within the MI framework. Many decision trees are trained using bootstrap samples in a random forest model. Estimates from from individual decision tree regressors are averaged to yield the final estimates. The split on feature values at each node of the decision tree is chosen optimally based on criteria derived from information theory. The gradient boosting trees algorithm uses a sequence of weak decision trees to learn the input-output mapping (regression or classification) and pass the error gradient to the following trees to boost the overall mapping accuracy. If $e_i$ is the error of the first decision tree to map input X to output y (Equation 14), the next tree in the sequence will learn $e_i$ from X and yield an error $e_{i+1}$ (Equation 15). Recursively, the output y is learned from Equation 16. 
\begin{eqnarray}
y&=&a_i+(b_i*X)+e_i \\
e_i&=&a_{i+1}+(b_{i+1}*X)+e_{i+1}\\
y &=& \sum_{i=1}^n a_i + X * \sum_{i=1}^n b_i + e_n
\end{eqnarray}
\subsection {Imputation aided by clustering}
We hypothesize in Section II-E that the knowledge of sample clusters can aid more accurate estimates of missing values being treated by homogeneity of data samples. The data set with missing values is first completed with medians of individual variables to enable clustering. A clustering algorithm identifies each sample ($x_i \in \Re^d$) belonging to one of $k$ disjoint groups of samples as $c_i \in \{c_1, c_2, ..., c_k \}$ with a cluster centroid, $\bar {x}_i \in \{\bar {x}_1 , \bar{x}_2, ..., \bar{x}_k \}$. A distance-based clustering algorithm updates the cluster centroids to minimize the sum of squared distance (SSE) between the samples within a cluster and the cluster centroid as below.
\begin{equation}
    SSE = \sum_{i=1}^{n} \sum_{p=1}^{k} w_{ip}|| x_i - \bar{x}_p||_2^2,
\end{equation}
where $\bar{x}_p$ is the centroid of the p-th cluster. The association between the i-th sample with the p-th cluster is
\begin{equation}
    w_{ip} = \begin{cases}
  1, \mbox {if $x_i$ belongs to the p-th cluster} \\    
  0, \mbox {otherwise.}    
\end{cases}
\end{equation}
Thus, the data set is clustered into $k$ groups labeling each sample under one cluster. The cluster label is infused into the training data samples in three ways, as shown in Algorithm 3. First, the cluster labels are appended as a new feature column to the samples as $X = \{x_i, c_i\}$. Second, one-hot encoded labels are appended as a vector of features, $X = \{x_i, e_{1i}, e_{2i},..., e_{ni}\}$, where n = ceil $(log_2^k)$. Third, the magnitude of cluster mean vector (MCMV) is proposed as a feature column, $X = \{x_i,~\bar {x}_i^T\bar {x}_i\}$. Thus, each variable ($x_j$) is also regressed from sample cluster information to introduce more homogeneity in MVI. The revised regression model for variable $x_j$, following Equation 4, becomes as below.
\begin{equation}
 x_j = f (x_k, c_i, W, ),~k \neq j.
\end{equation}
In other words, the regression model for variable $x_j$ is now a function of other variables $x_k$ and the cluster variable $c_i \in \{c_1, c_2, ..., c_k \}$ corresponding to the data sample $i$, weighted by the model parameter vector W. 
\begin{algorithm}[t]
\caption{Ensemble learning and clustering within MI}
\begin{algorithmic}
\STATE Input: Data matrix $X = \{X_{obs}, X_{miss}\}\in\Re^{n\times d}$ = $\{x_i\}$
\STATE Output: Imputed data matrix, $X' = \{X_{obs}, X_{imputed}\}$
\STATE $X_{median}$ = median ($X_{obs}$)
\STATE $X_{temp} \leftarrow \{X_{obs}$, impute $X_{miss}$ with $X_{median}$\}
\STATE Obtain $k$ clusters \{$C_t$\} in $X_{temp}$, t = 1, 2, ...,k
\STATE C  $\leftarrow$ cluster labels for samples  
\IF {method == 'label'}
\STATE $X_G$ = concatenate (X, C)
\ELSIF {method == 'one-hot'}
\STATE $e_1, e_2, .., e_n$ = one-hot~(C)
\STATE $X_G$ = concatenate (X, $e1, e2, .., e_n$)
\ELSIF {method == 'MCMV'}
\FOR {p = 1 $\rightarrow$ k}
\STATE $X_p$ = $X_{temp}$ ($C_t$ == p)
\STATE cluster mean vector, $C_{mean}$ = $\frac{1}{N_p}\sum_{i=1}^{N_p} X_p (i)$
\STATE $G_p$ = $C_{mean}^T . C_{mean}$
\ENDFOR
\STATE $X_G$ = concatenate (X, $G_p$)
\ENDIF
\STATE Under multiple imputations loop of Algorithm 1 
\FOR {t = 1 $\rightarrow$ $N$}
\STATE $X_{imputed}^t$ = Ensemble$\_$MICE ($X_G$)
\STATE $X^t \leftarrow \{X_{obs}, X_{imputed}^t \}$
\STATE Update $X_G$ $\leftarrow$ $X^k$
\STATE Continue until $|X_{imputed}^{t+1} - X_{imputed}^t|\leq \tau$
\ENDFOR
\RETURN $X' = \{X_{obs}, X^{N}_{imputed}\}$
\end{algorithmic}
\end{algorithm}

\begin{figure}[t]
\vspace{0pt}
	\includegraphics [width=9cm]{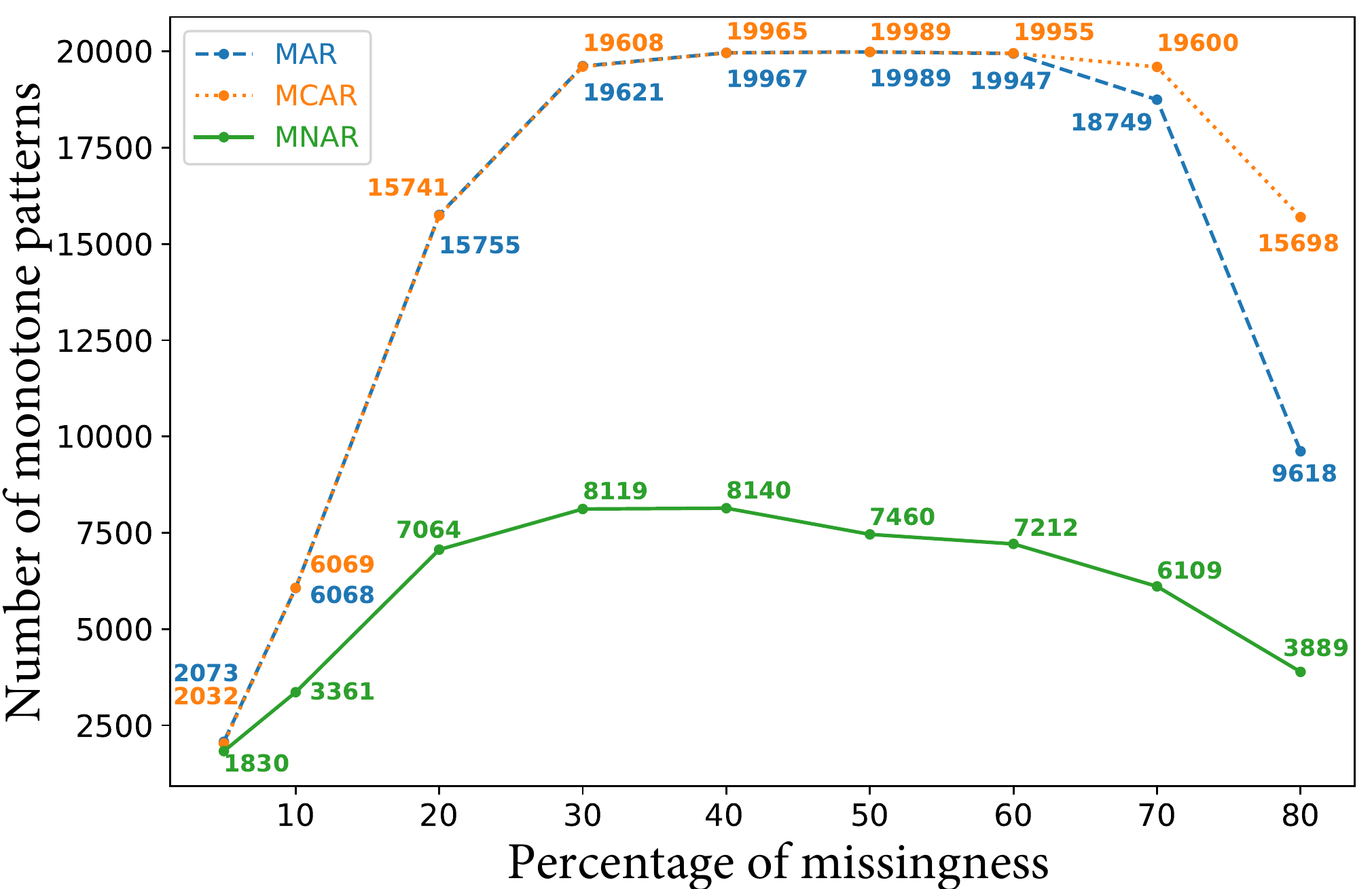}
	\vspace{-10pt}
	\caption{The number of monotone patterns versus the percentage of missing values for MAR, MCAR and MNAR type of missingness for default of credit card clients data set. }
	\label{figMonotone}
\end{figure}
\begin{figure*}[t]
\centering
\subfigure[Breast cancer] { \includegraphics[trim=.6cm .6cm .6cm .6cm, width=0.32\textwidth] {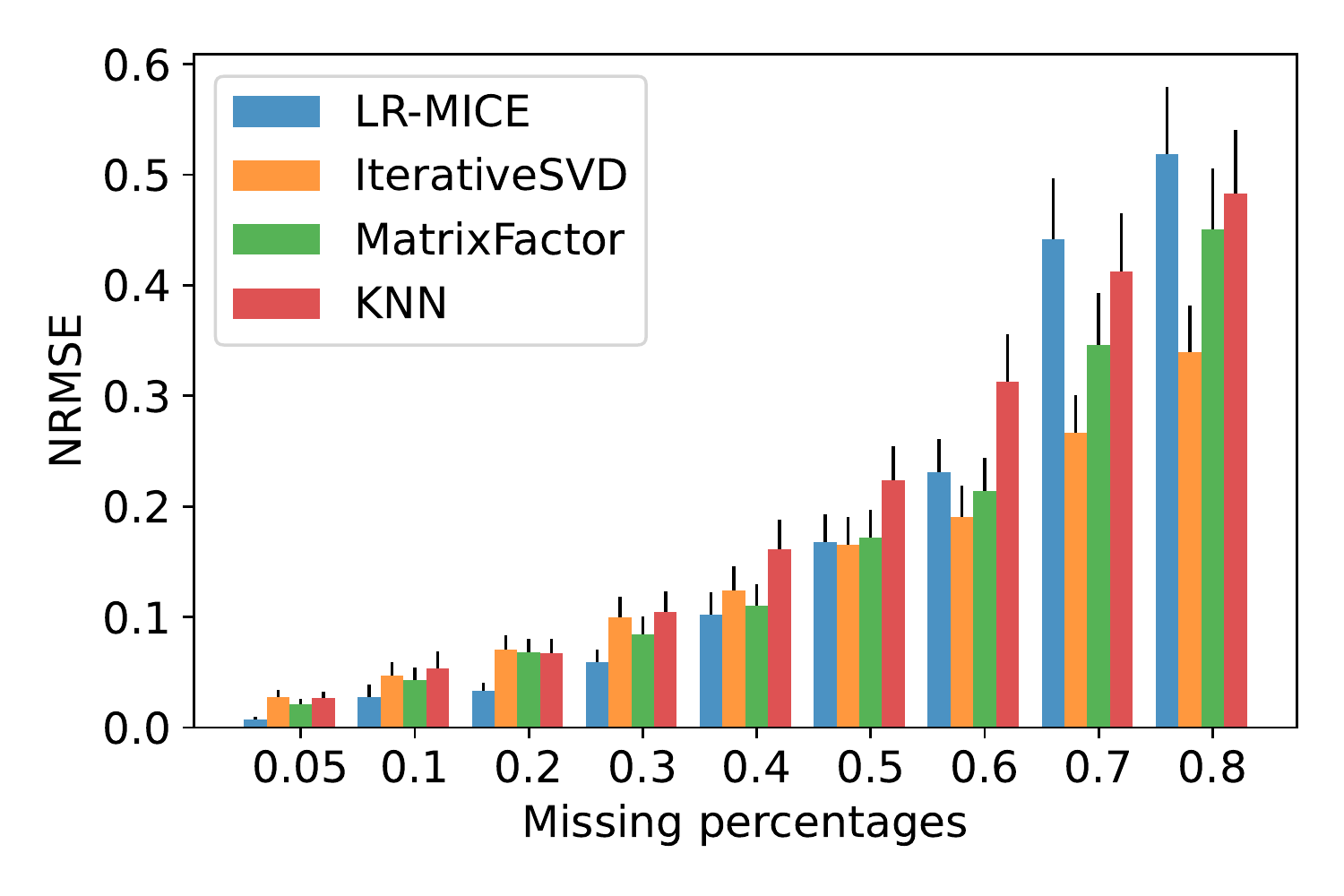}}
\subfigure[Dermatology] { \includegraphics[trim=.6cm .6cm .6cm .6cm, width=0.32\textwidth]  {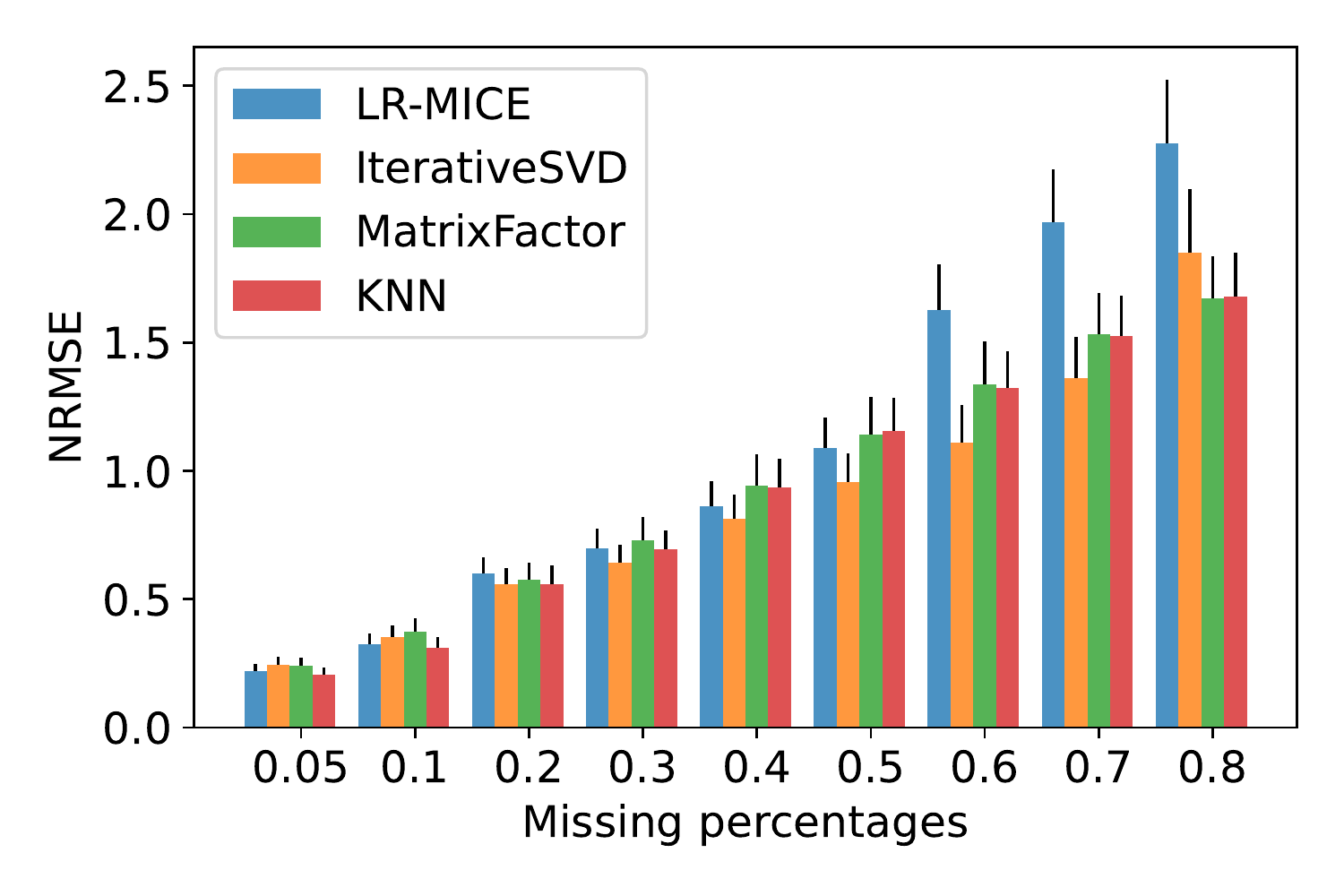}}
\subfigure[skillcraft1] { \includegraphics[trim=.6cm .6cm .6cm .6cm, width=0.32\textwidth] {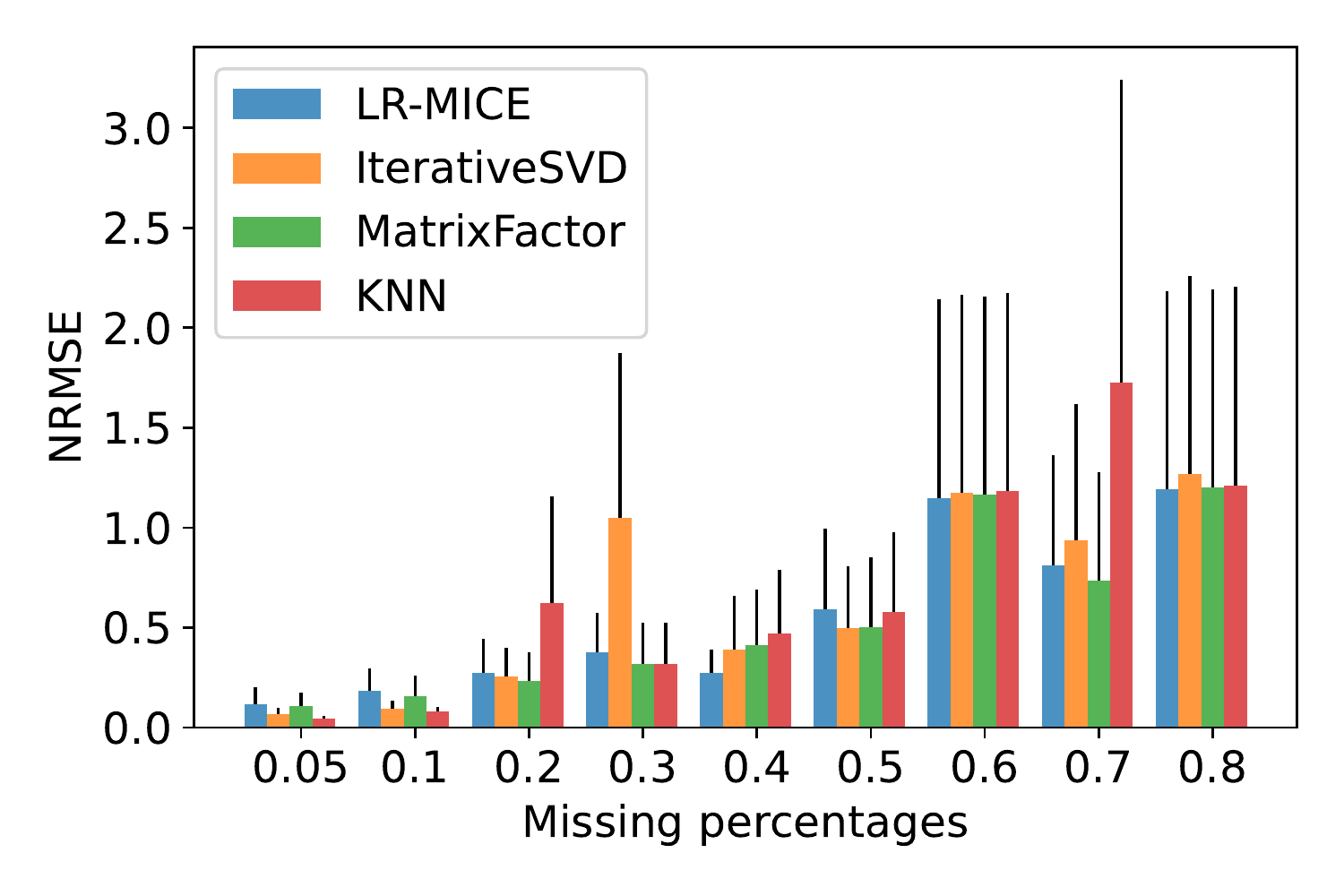}}
\subfigure[Wine quality] { \includegraphics[trim=.6cm .6cm .6cm .6cm, width=0.32\textwidth]{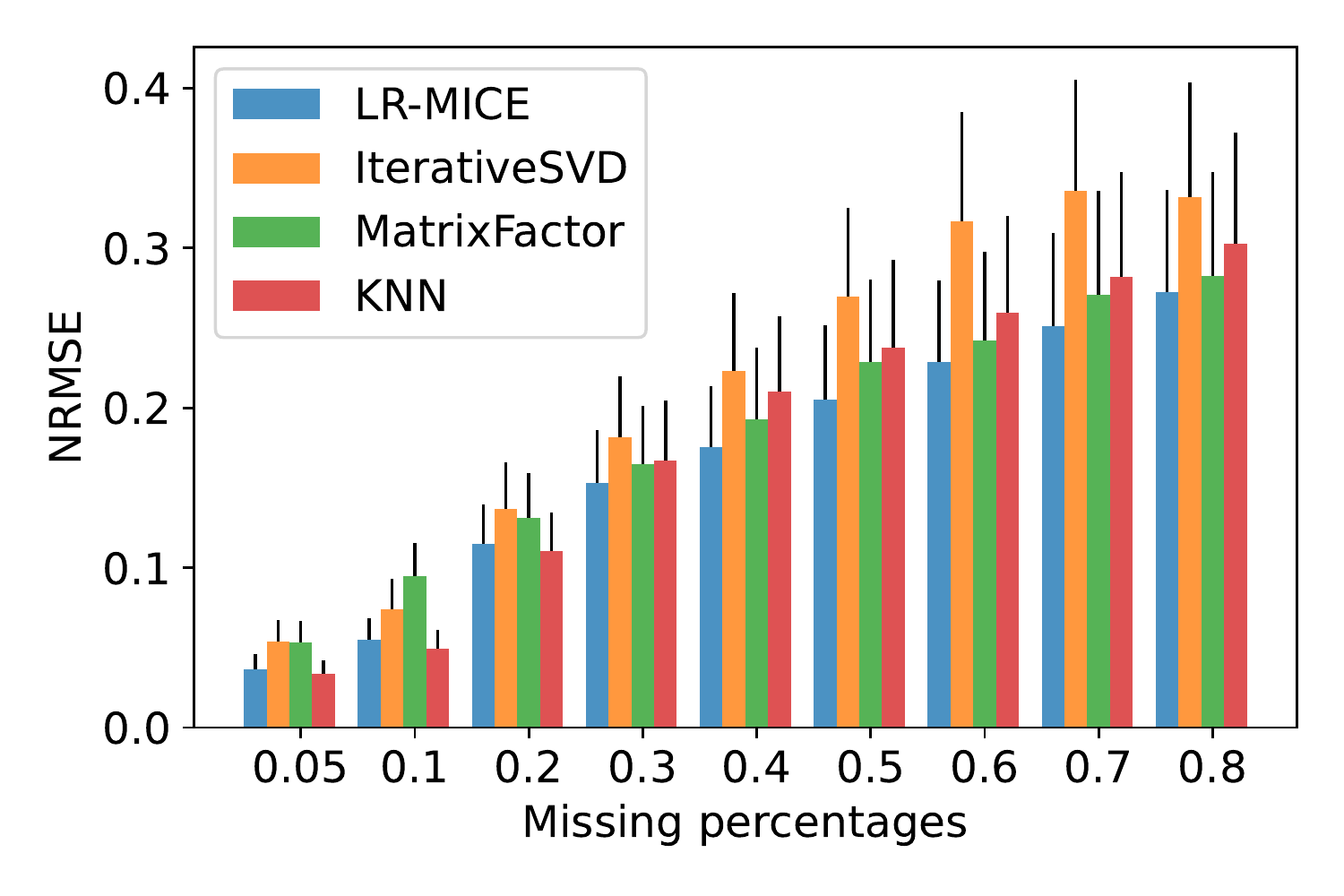}} 
\subfigure[Default of credit card clients] { \includegraphics[trim=.6cm .6cm .6cm .6cm, width=0.32\textwidth] {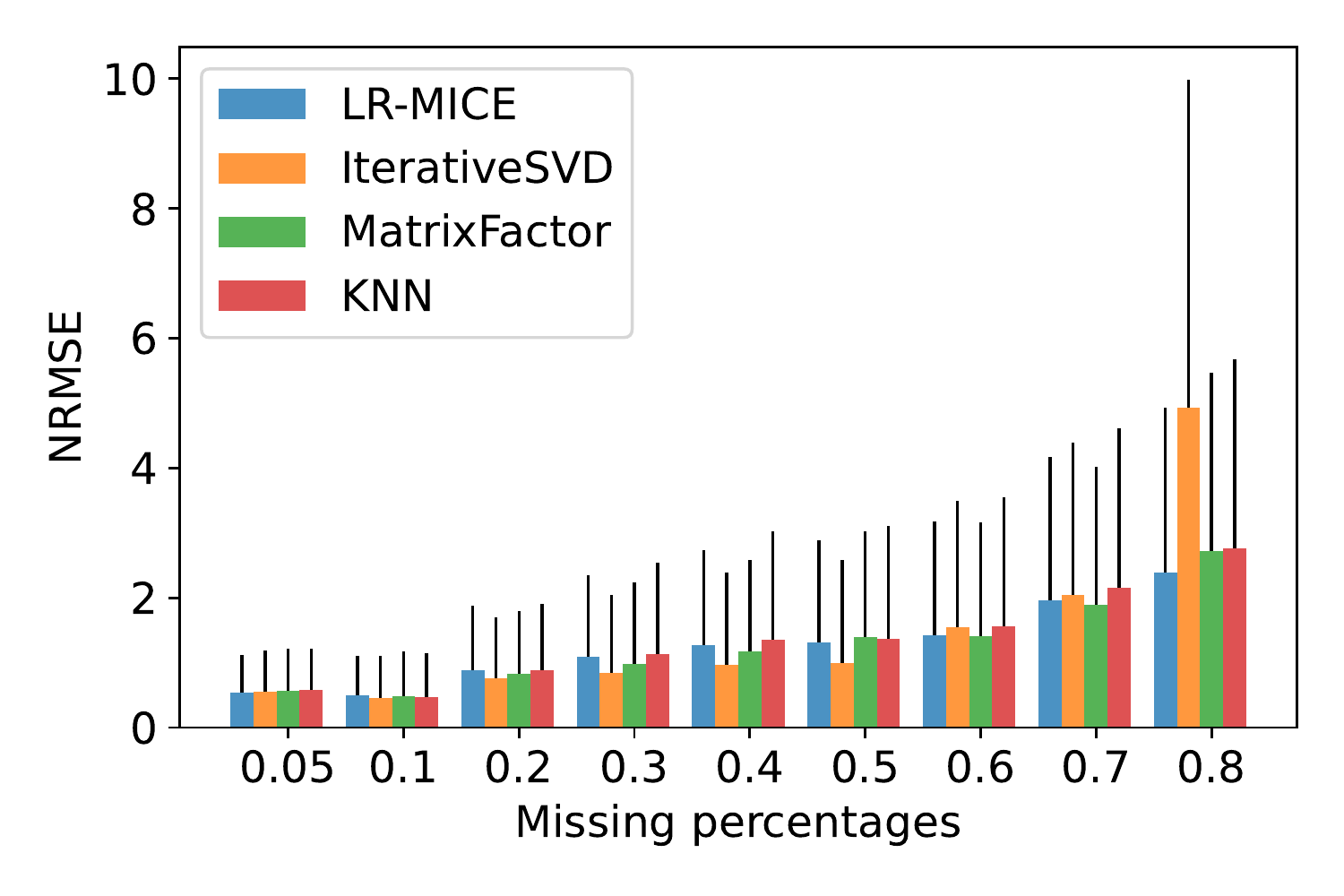}}
\subfigure[Mice protein expression] { \includegraphics[trim=.6cm .6cm .6cm .6cm, width=0.32\textwidth]  {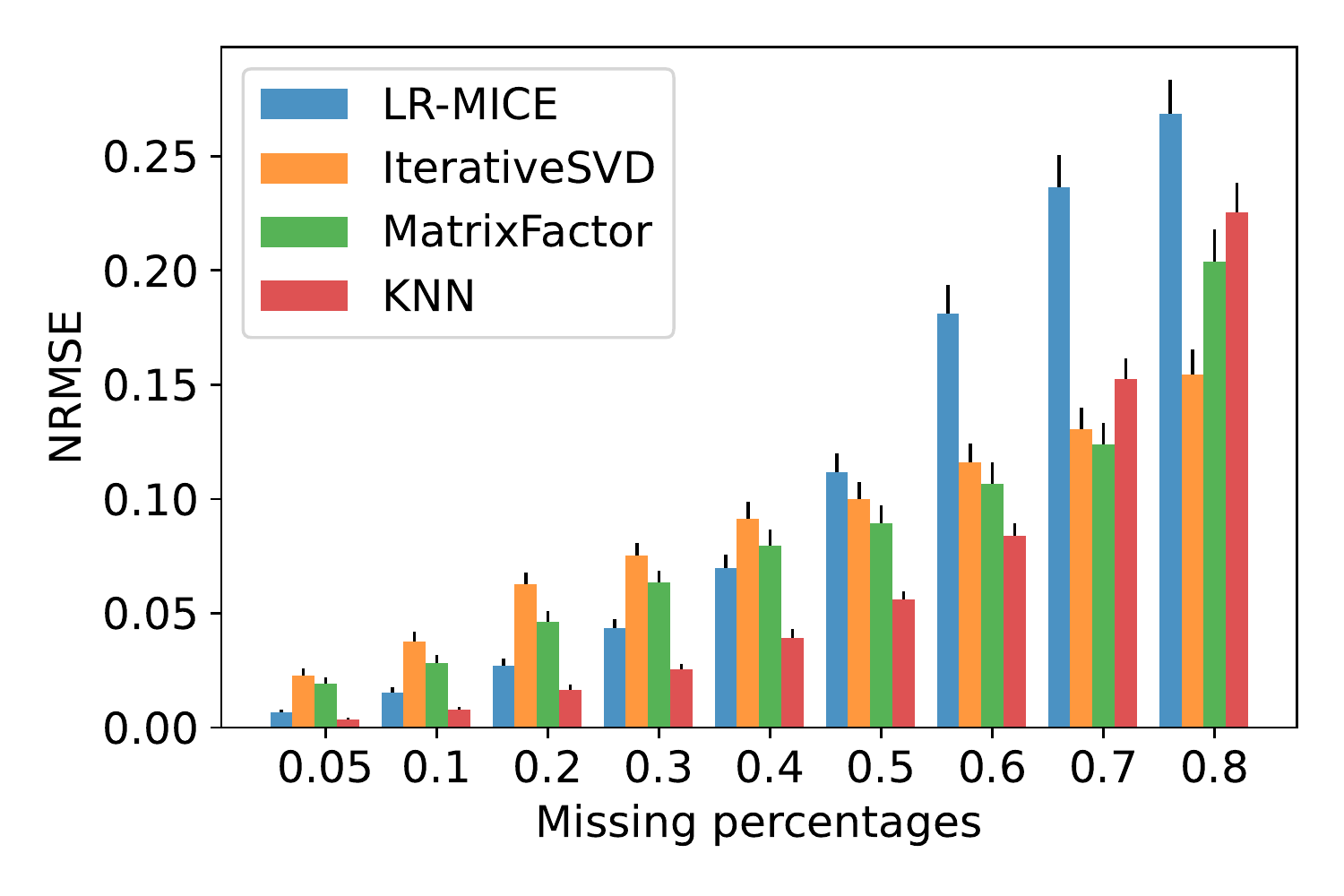}}
\caption{Normalized root mean squared error (NRMSE) comparison between MICE (linear regression-MICE or LR-MICE) and non-MICE algorithms  (iterative SVD, matrix factorization, K-nearest neighbor (KNN)). The test data are missing at random (MAR).}
\label{figNRMSE}
\end{figure*}
\subsection {Experimental design}
The imputation accuracies of the baseline and proposed algorithms are obtained and compared through extensive experimentation. First, the robustness of the algorithms is evaluated for data sets with 5\% to 80\% missing values. It should be noted that only 10.8\% of the literature have performed imputation for more than 50\% missing values~\citep{Lin2020}. Second, we compare five categories of imputation algorithms: 1) non-MICE imputations (KNN, iterative SVD, matrix factorization) using the fancyImpute package~\citep{fancyimpute2016}, 2) the baseline MICE algorithm (MI with chains of linear regression), 3) MI with chains of ensemble learning (gradient boosting and random forest), 4) MI with chains of deep regressors, and 5) MI with ensemble learning plus the cluster information. The first two categories will serve as the baseline models. Third, the experimentation will be conducted for imputing three missingness types (MAR, MCAR, and MNAR) across data sets of different domains.

\subsection {Model evaluation}
The imputation of missing values is performed on standardized data set with zero mean and unit standard deviation. Following the imputation, the data set is transformed back to its original scale to compare the imputed values against their actual counterparts. The performance of the imputation models is evaluated in two ways. First, the trained imputation model imputes a separate test data set with missing values. The error between imputed and actual values on the test data set is measured by the root mean squared error (RMSE) (Equation 20) divided by the mean of the actual values as the normalized root mean square error (NRMSE) (Equation 21). 
\begin{eqnarray}
RMSE &=& \sqrt{\frac{\sum_{i=1}^{N}(X_i - \hat{X_i})^2}{N}}\\
NRMSE &=& \frac{RMSE}{\frac{1}{N}\sum_{i=1}^{N}X_i}
\end{eqnarray}
Second, the quality of imputed data is evaluated in supervised classification tasks using the random forest classifier. A 10x2 fold nested cross-validation is performed first to search the best model hyperparameters (two inner folds) and then report the mean classification accuracy across the test folds (ten outer folds). The outcomes of the two evaluation approaches will reveal if better imputation accuracy translates to better classification accuracy.  
\begin{table*}[t]
\caption{Best MICE models for each data set after voting across all missingness percentages.}
\begin{tabular}{@{}llll@{}}
\toprule
\multirow{2}{*}{Datasets}        & \multicolumn{3}{c}{Best Model}                              \\ \cmidrule(l){2-4} 
                                 & MAR                & MCAR                & MNAR               \\ \cmidrule(r){1-4}
Breast cancer         & LR-MICE + Label    & LR-MICE + One-hot   & LR-MICE + One-hot  \\
Dermatology                      & RF-MICE + One-hot  & RF-MICE + One-hot   & RF-MICE + One-hot  \\
SkillCraft1 master          & DR-MICE            & DR-MICE            & DR-MICE           \\
Wine quality                     & GB-MICE + Label    & GB-MICE + MCMV  & LR-MICE + One-hot  \\
Default of credit card   & RF-MICE    & GB-MICE + One-hot  & GB-MICE + One-hot            \\
Mice protein   expression        & LR-MICE + Label    & RF-MICE + MCMV  & RF-MICE + MCMV \\ \bottomrule
\end{tabular}
\label{tab02}
\end{table*}
\begin{table*}[t]
\caption{Comparison of NRMSE scores of proposed inclusion of cluster information (label, one-hot, MCMV) with the baseline MICE (LR-MICE) on the wine data set. }
\resizebox{\textwidth}{!}{
\begin{tabular}{@{}cllllllllllll@{}}
\toprule
& \multicolumn{4}{c}{MAR}                                                                                                                         & \multicolumn{4}{c}{MCAR}                                                                                                                        & \multicolumn{4}{c}{MNAR}                                                                              \\ \midrule
\begin{tabular}[c]{@{}l@{}}Missing   \\\end{tabular} & LR-MICE       & \begin{tabular}[c]{@{}l@{}}LR-MICE \\ + Label\end{tabular} & \begin{tabular}[c]{@{}l@{}}LR-MICE \\ + One-hot\end{tabular} & \begin{tabular}[c]{@{}l@{}}LR-MICE \\ + MCMV\end{tabular} & LR-MICE                & \begin{tabular}[c]{@{}l@{}}LR-MICE\\ + Label\end{tabular} & \begin{tabular}[c]{@{}l@{}}LR-MICE\\ + One-hot\end{tabular} & \begin{tabular}[c]{@{}l@{}}LR-MICE \\ + MCMV\end{tabular} & LR-MICE       & \begin{tabular}[c]{@{}l@{}}LR-MICE\\ + Label\end{tabular} & \begin{tabular}[c]{@{}l@{}}LR-MICE\\ + One-hot\end{tabular} & \begin{tabular}[c]{@{}l@{}}LR-MICE\\  + MCMV\end{tabular} \\ \midrule
5\%                                                             & 0.0377  & \textbf{0.0367 }                                     & 0.0369                                                & 0.0368                                                  & \textbf{0.0397 } & 0.0398                                              & 0.0402                                                & 0.0399                                                  & 0.146   & 0.1452                                            & \textbf{0.1445 }                                      & 0.1449                                                  \\
10\%                                                            & 0.0551  & 0.0555                                            & \textbf{0.0548 }                                       & 0.055                                                  & 0.0595           & 0.0587                                            & \textbf{0.0577 }                                      & 0.0589                                                 & 0.1754  & 0.1746                                            & \textbf{0.174 }                                       & 0.1746                                                 \\
20\%                                                            & 0.1147  & \textbf{0.1136 }                                     & 0.1136                                                & 0.1147                                                & 0.1125          & 0.1119                                              & \textbf{0.111 }                                       & 0.1119                                                 & 0.2164  & 0.2153                                             & \textbf{0.215 }                                       & 0.2157                                                  \\
30\%                                                            & 0.1529  & 0.1515                                             & \textbf{0.1515 }                                       & 0.1519                                                  & 0.1486          & 0.1529                                              & 0.1479                                              & \textbf{0.1468 }                                        & 0.2417  & 0.2411                                              & \textbf{0.2409 }                                      & 0.2412                                                 \\
40\%                                                            & 0.1754  & 0.1756                                             & \textbf{0.1748 }                                       & 0.1748                                                & 0.1823           & 0.1814                                            & \textbf{0.1809 }                                      & 0.182                                                 & 0.2655  & 0.2652                                              & \textbf{0.2652 }                                      & 0.2656                                                 \\
50\%                                                            & 0.2051  & 0.2043                                              & \textbf{0.2032 }                                       & 0.2043                                                  & 0.2032           & 0.2031                                             & \textbf{0.2021 }                                      & 0.2028                                                 & 0.2772  & 0.2766                                             & \textbf{0.2766 }                                      & 0.2767                                                  \\
60\%                                                            & 0.2289  & 0.2292                                               & \textbf{0.2264 }                                       & 0.2274                                                  & 0.2217          & 0.2244                                             & 0.2227                                                & \textbf{0.2214 }                                        & 0.2883  & 0.2874                                            & 0.2874                                               & \textbf{0.2871 }                                        \\
70\%                                                            & 0.2513  & 0.2567                                             & 0.2569                                                 & \textbf{0.2488 }                                        & \textbf{0.2496 } & 0.256                                               & 0.2578                                                & 0.2504                                                & 0.2958  & 0.2957                                           & \textbf{0.2956 }                                      & 0.2957                                                  \\
80\%                                                            & 0.2722  & 0.2724                                            & 0.2724                                                 & \textbf{0.2716 }                                        & 0.272           & 0.2802                                             & 0.2795                                               & \textbf{0.2708 }                                        & 0.2993  & 0.2992                                             & \textbf{0.2992 }                                      & 0.2993  \\ \bottomrule                                            
\end{tabular}}
\label{tab03}
\end{table*}

\section {Results}
The proposed experiments are executed on a Dell Precision 5820 workstation running Ubuntu 20.04 with 64GB RAM and an Nvidia Quadro 5000 GPU with 16GB memory. All algorithms and evaluation steps are implemented in Python. The findings of the proposed simulation and experimentation are discussed below.

\subsection {Simulating missing values}

We simulate three missingness types following Algorithm 2. The simulation of missingness is evaluated by counting monotone patterns in missing values as a function of the percentage of missingness~\citep{VanStein2016}. More monotone patterns indicate that the missingness in data has more more uniquely and randomly distributed patterns with minimum matching. A low number of monotone patterns indicates more matching or similar patterns in missingness. Figure~\ref{figMonotone} shows that MNAR has the least number of monotone patterns suggesting minimal randomness in the missing value pattern. On the other hand, MCAR has more monotone patterns than MAR, as observed in data with 60\% or more missing values. The number of monotone patterns (or the number of unique patterns) in missing values peaks between 30\% and 50\% data missingness.

\subsection {Linear MICE versus non-MICE models}
We compare the popular MICE model (LR-MICE) with other non-MICE imputation methods, including iterative singular value decomposition (Iterative SVD), matrix factorization, and K-nearest neighbor (KNN). Figure~\ref{figNRMSE} shows the effect of the percentage of missingness on NRMSE scores for the MAR type data. Intuitively, all data sets show a linear relationship between the NRMSE scores and the percentage of missing values. A general observation is that the performance of the MVI algorithms depends on the data set. Therefore, selecting one MVI model as the best for all datasets is likely to yield suboptimal results.  The LR-MICE model generally performs well at a lower percentage of missing values ($<30\%$). LR-MICE appears to be the best for the the wine quality data set at any percentages of missingness. 
The KNN algorithm outperforms LR-MICE up to 60\% missingness in the mice protein expression data set. The iterative SVD algorithm yields the best NRMSE scores at higher missing percentages ($>$50\%), especially with the breast cancer and dermatology datasets.  These observations contradict the general notion that the baseline MICE is the best model for imputing missing values in tabular data.



\begin{table*}[t]
\caption{The accuracy of a random forest classifier in classifying the dermatology data set imputed by the gradient boosting (GB)-MICE algorithms. The mean classification accuracy is obtained across ten cross-validation folds. The accuracy on the actual dermatology data set (without missing values) is 98.04\%.}
\begin{tabular}{ccccccc}
\toprule
\multicolumn{5} {c}{MAR} & MCAR & MNAR \\
\cmidrule{2-5}
Missing              & GB-MICE & GB-MICE                  & GB-MICE                   & GB-MICE&GB-MICE&GB-MICE\\
           & (baseline)        &+ Label                                &+ One-hot                              &+ MCMV&+ MCMV&   + MCMV\\
\midrule
5\%                    & 0.964 (0.03)      & 0.969 (0.03)                          & 0.967 (0.03)                          & \textbf{0.969 (0.02)}&0.975 (0.03) &0.967(0.03) \\
10\%                   & 0.969 (0.02)      & 0.969 (0.02)                          & 0.972 (0.02)                          & \textbf{0.975 (0.02)} &0.978 (0.02)&0.955 (0.02)\\
20\%                   & 0.972 (0.02)      & \textbf{0.978 (0.02)}                 & 0.972 (0.02)                          & 0.969 (0.02)       &0.975 (0.04)&0.927 (0.03) \\
30\%                  & 0.955 (0.02)      & 0.953 (0.02)                          & 0.958 (0.02)                          & \textbf{0.964 (0.02)} &0.958 (0.02)&0.899 (0.03)\\
40\%                   & 0.930 (0.04)      & 0.922 (0.04)                          & 0.936 (0.05)                          & \textbf{0.939 (0.04)} &0.922 (0.02)&0.804 (0.02)\\
50\%                   & 0.924 (0.07)      & \textbf{0.927 (0.05)}                 & 0.927 (0.05)                          & 0.927 (0.06)        &0.916 (0.02)&0.793 (0.04)\\
60\%                   & 0.846 (0.06)      & 0.857 (0.04)                          & 0.852 (0.06)                          & \textbf{0.863 (0.07)}&0.874 (0.01)&0.752 (0.03) \\
70\%                   & 0.757 (0.06)      & 0.779 (0.06)                          & \textbf{0.782 (0.06)}                 & 0.751 (0.03)      &0.805 (0.06)&0.470 (0.08) \\
80\%                   & 0.662 (0.09)      & 0.676 (0.10)                          & 0.657 (0.08)                         & \textbf{0.696 (0.08)}&0.670 (0.05)&0.327 (0.07)\\
\bottomrule
\end{tabular}
\label{tab04}
\end{table*}
\begin{table*}[t]
\caption{The best missing value imputation (MVI) algorithm for each data set and missingness type pair. The values represent the lowest sum of NRMSEs (therefore the best model) across all missing percentages. The best model is chosen from six baseline MVI models (Iterative SVD, matrix factorization, KNN, LR-MICE, RF-MICE, GB-MICE) and three proposed models (RF-MICE-Cluster, GB-MICE-Cluster, and DR-MICE). Here, the 'Cluster' represents cluster labels (label), one-hot encoded labels (one-hot), and the magnitude of cluster mean vector (MCMV).}
\begin{tabular}{@{}lllllll@{}}
\toprule
\multirow{2}{*}{Data sets}     & \multicolumn{3}{c}{Best model}                          & \multicolumn{3}{c}{Lowest sum of NRMSE} \\ \cmidrule(l){2-7} 
                               & MAR                & MCAR              & MNAR              & MAR         & MCAR        & MNAR        \\ \cmidrule(r){1-7}
Breast cancer        & Iterative SVD      & RF-MICE + MCMV     & Iterative SVD     & 1.33      & 1.24      & 2.77      \\
Dermatology                    & RF-MICE + One-hot  & GB-MICE + One-hot  & Iterative SVD     & 7.47      & 7.36      & 9.70      \\
SkillCraft1       & DR-MICE            & DR-MICE       & DR-MICE          & 3.32      & 3.38      & 9.57      \\
Wine quality                   & GB-MICE + MCMV     & GB-MICE + One-hot   & LR-MICE + One-hot & 1.47      & 1.47      & 2.20      \\
Default of credit & RF-MICE + One-hot     & GB-MICE + One-hot  & LR-MICE       & 9.36      & 16.47      & 19.12     \\
Mice protein        & DR-MICE            & KNN               & DR-MICE           & 0.59      & 0.56      & 1.29      \\ \bottomrule
\end{tabular}
\label{tab05}
\end{table*}

\subsection{Machine learning in MICE}

We replace the linear equations of the baseline MICE model with non-linear machine learning regression models, including decision tree (DT), random forest (RF), gradient boosting (GB) trees, and deep regression (DR) with three hidden layers. The number of neurons in the first hidden layer is set based on the number of input variables. Each successive layer has half the number of neurons compared to the previous layer. Understandably, the DT-MICE model performs the worst among all non-linear MICE models because of not being an ensemble learning with a single decision tree. The MVI methods with ensemble learning in the MI framework (GB-MICE and RF-MICE) appear most frequently in the best model chart compared to the baseline LR-MICE model (Table~\ref{tab02}). 

The MCAR type mostly benefits from ensemble learning within the MI framework (RF-MICE, GB-MICE). This benefit appears to be the least in imputing data with the MNAR type. Interestingly, the skillcraft1 data set yields the best imputation results with the DR-MICE imputation model regardless of the missing value types, as shown in Table~\ref{tab02}.  However, the DR-MICE model takes a much higher computing time compared to other models. For example, our workstation takes six hours on average to complete GB-MICE imputation for all nine missing percentages. In contrast, DR-MICE takes approximately 48 hours to impute the same data, which is eight times more. In almost all cases, the inclusion of cluster information (one-hot encoding or labels or MCMV) in the imputation process augments the performance of ensemble learning within the MI framework. The only exception is the default of credit data set with MAR values where RF-MICE outperforms those with the cluster information. 

Furthermore, we investigate the effect of including the proposed cluster information in the baseline LR-MICE model alone. The number of clusters is chosen to be three based on the silhouette scores. The comparison of NRMSE scores in Table~\ref{tab03} reveals that data with the MNAR type are almost always best imputed by LR-MICE plus one-hot encoded cluster information. The inclusion of cluster information to LR-MICE consistently outperforms the baseline LR-MICE for any missingness types or percentage of missing values. The only exceptions are found at 5\% and 70\% of the MCAR type, where the baseline LR-MICE outperforms the models with cluster information. However, these percentage values are less common, and so is the MCAR type in real-world data sets.

\subsection{Comparison of classification accuracy}

We investigate if improved data imputation also yields improved classification accuracy. That is, the imputation model yielding the lowest NRMSE score is expected to contribute to the best classification accuracy. Table~\ref{tab03} shows that MNAR is the most challenging type of missingness, resulting in the highest NRMSE values, which is followed by MAR and MCAR types. However, the differences between the missingness types diminish in classification accuracy. Table~\ref{tab04} shows the quality of GB-MICE imputed data in a six-class classification task using the dermatology data set. As expected, the classification accuracy obtained using imputed data sets never exceeds the one with actual or complete data set (98.04\%). However, the classification accuracy is substantially affected (drops below 90\%) when the data set is imputed for more than 50\% missing values (MCAR, MAR types) and more than 30\% missing values (MNAR type). The inclusion of data cluster information yields superior classification accuracy for all missing percentages than the baseline GB-MICE without the cluster information. Therefore, our proposed inclusion of data cluster information improves the MVI accuracy, which contributes to superior classification accuracy using the imputed data set. A data set, imputed with the aid of the cluster information represented by MCMV, yields the most robust improvement in the classification accuracy, as observed in Table~\ref{tab04}. It is noteworthy that the cluster information is only used during the imputation step and excluded at the time of classification. 

\subsection{Overall comparison of imputation models}
We finally compare the performance of ten MVI models: 1)the baseline LR-MICE, 2) LR-MICE with the cluster information, 3-5) non-MICE imputation algorithms (iterative SVD, matrix factorization, KNN),  6-7) ensemble learning within the MI framework (RF-MICE, GB-MICE), 8) deep learning within the MI framework (DR-MICE), and 9-10) MI with ensemble learning plus the cluster information (MI-EL-CI). Here, MI-EL-CI includes the cluster information in GB-MICE and RF-MICE models. Table~\ref{tab05} sums the NRMSE scores across all missing percentages for each imputation model. DR-MICE yields the best performance when imputing the skillcraft1 data set regardless of missingness types.

Three out of six data sets with MAR values show superior performance with our proposed MI-EL-CI imputation algorithms. The proposed DR-MICE performs the best for two other data sets with MAR. Five out of six data sets with MCAR values find our proposed MI-EL-CI or DR-MICE imputation superior to all other algorithms. Conversely, the imputation of data MNAR type yields mixed findings. In general, ensemble learning within MI does not perform the best in imputing any of the six datasets with MNAR values. Iterative SVD, LR-MICE, and our proposed DR-MICE individually have ranked the best in imputing two datasets with MNAR. The NRMSE values in Table~\ref{tab05} once again confirm that imputing MNAR is the most challenging type with the highest errors among the three missingness types.


\section{Discussion}

This paper aims to improve the performance of the baseline MICE model by incorporating deep learning and cluster information. Extensive evaluations of the proposed models across six data sets, three missingness types, and up to 80\% missingness have revealed several important findings as follows. First, ensemble learning, replacing the linear regression equations in the MICE framework, improves the MVI accuracy for MAR and MCAR types (Tables~\ref{tab02},~\ref{tab05}) but not for the MNAR type. Second, the inclusion of cluster labels almost always improves the imputation accuracy of ensemble-based MICE (Table~\ref{tab02}) and baseline MICE (LR-MICE) models (Table~\ref{tab03}). This observation is across all percentages of missingness. Third, the inclusion of cluster labels in MVI yields datasets that achieve a better classification accuracy compared to the data set imputed without the cluster information (Table~\ref{tab04}). The improvement in classification accuracy is consistently observed in one of the best MICE model with ensemble learning (GB-MICE). Fourth, missing not at random (MNAR) is the most challenging among the three types of missingness, resulting in the highest NRMSE and lowest classification accuracy (Tables~\ref{tab03}-\ref{tab05}). The best MVI and classification accuracies are observed in the MCAR type, followed by the MAR type. Fifth, both MAR and MCAR types are best imputed by our proposed ensemble plus cluster-based MICE and DR-MICE models (Table~\ref{tab05}). In contrast, the MNAR type is better imputed with iterative SVD and DR-MICE than ensemble-based MICE models. Sixth, MICE-based MVI models are superior to non-MICE models with up to 30\%-50\% missing data. Non-MICE MVI models, such as iterative SVD and KNN, have been found to outperform the baseline MICE model at higher missingness percentages ($>$50\%). (Figure~\ref{figNRMSE}).

\subsection {Effects of missingness types}

The missingness type in data manifests several notable effects on the performance of MVI algorithms. Figure~\ref{figMonotone} shows the varying number of monotone patterns across different percentages of missingness for different missingness types. This variability in monotone patterns may have impacted the MVI performance. According to~\citep{Lin2020}, only 15 out of their 111 surveyed papers (13.5\%) have studied all three missingness types, including MNAR. Therefore, MVI models in the literature are often limited to only MCAR type, which may not perform equally well on more realistic types of missingness such as MAR and MNAR as observed in our study. It is noteworthy that the MAR type is defined under the same conditional dependence on observed data that is used to formulate MICE-based imputation models. Therefore, it is no wonder that MICE will yield superior performance with MAR or MCAR types. However, the superiority of the MICE framework in previous studies may not be generalized for the MNAR type. In MNAR, missingness depends on the missing values rather than the observed part of the data. Therefore, MICE-based models can yield suboptimal performance because missing values do not depend on the observed values in the MNAR type. Our results consistently show that the imputation and classification accuracies for the MNAR type are substantially more vulnerable to increasing percentages of missingness than MAR or MCAR types.

\subsection {Effects of clustering on imputation}

The addition of cluster information to individual data samples has consistently improved the MVI accuracy over the baseline MICE and ensemble learning in MICE models. That is, a missing value is regressed from its sample cluster information in addition to other observed variables. In standard MICE, a missing value is regressed from heterogeneous samples that can be widely different from the sample under imputation. The chains of equations in MICE remain indifferent about the contribution of heterogeneous data samples, which may introduce high variability in estimating the missing values. Including the cluster information in data treats samples of the same cluster more equally than those from different clusters, improving the imputation accuracy. The improved imputation accuracy has also resulted in better quality data for training and testing classification models. 

\subsection {Effects of data sets}

The performance of the baseline and proposed MVI algorithms has revealed some dataset-specific trends. It is obvious that not all data sets are imputed equally well by one MVI algorithm. Therefore, papers reporting imputation accuracy on a single data set or domain data may not generalize well on other data domains. The wine quality data set shows the most robust performance of the baseline MICE model compared to non-MICE models in Figure~\ref{figNRMSE}(d), even for a high percentage of missingness in data. In contrast, the MICE model yields the worst performance at high missingness percentages ($>50\%$) when imputing the mice protein expression data set. It is noteworthy that the wine quality and mice datasets have the lowest (11) and highest (77) number of variables. Intuitively, a large number of variables may complicate the performance of the MICE model because of involving one regression model per variable. The most conspicuous discrepancy is noted with the skillcraft1 data set in Figure~\ref{figNRMSE}(c) at varying percentages of missingness. The skillcraft1 data set is invariably best imputed by DR-MICE regardless of the missingness types (Tables~\ref{tab02},\ref{tab05}). It appears that deep neural networks in the MI framework are robust for a particular data set (skillcraft1) and also in imputing datasets with MNAR type. Unlike other data sets, the skillcraft1 data set contains all positive values within a large range between zero and one million. Eight out of 19 variables of this data set have mean values less than 0.01, starting at the third decimal point. Data sets with similar distribution and statistics may be better imputed by the DR-MICE model. The DR-MICE model also performs the best in imputing the mice protein expression data set with the highest number of variables (Table~\ref{tab05}). The deep neural network sequentially down samples the data dimensionality by half at each of the three hidden layers, which may benefit the imputation of high dimensional data.

\subsection {Selection of imputation models}

The baseline MICE (LR-MICE) is often compared against any newly proposed MVI algorithms in the literature~\citep{neuroISP, Awan_Elsevier_2021}. We improve the MICE model by incorporating more accurate ensemble and deep regressors, which can be taken for future model comparisons. The concept of ensemble learning in the MI framework appears in some sporadic papers on specific data domains, which are yet to receive attention similar to the baseline MICE model. Our extensive analyses on both MICE and non-MICE MVI algorithms provide several recommendations considering missingness types, missingness percentages, and data domains. 

The baseline MICE model appears to be effective in most MVI studies with missingness within 30\%. Our results reveal that the baseline MICE model performance substantially deteriorates for data with more than 50\%  missing values compared to non-MICE models. In two extreme cases: 1) high missingness percentages ($>$50\%) and 2) MNAR type, we identify that non-MICE algorithms, such as iterative SVD and KNN, outperform MICE-based models. This observation is explainable because the number of observed values also drops with increasing missing data, whereas the MNAR type does not depend on observed values. In contrast, the chains of equations in the MICE model are built by conditioning on observed values. Furthermore, with increasing missingness, the data matrix is proportionally filled with more default values (e.g., median) to initiate an MVI algorithm. The presence of default values results in low-rank data matrices. Therefore, low-rank matrix decomposition techniques, such as SVD, may more effectively estimate missing values in such data.   

\subsection {Limitations}

Despite promising results, this work has several limitations, like any other study. Our proposed models are susceptible to a higher percentage of missingness and the MNAR type, which is an inherent limitation of the MICE framework. Despite the promising performance, the proposed DR-MICE model takes excessive time to train, losing its practical value. Our models are built on regressors, which are suitable for real-valued data, not categorical variables. The performance of MVI models does not explain how these models correlate with different data distributions, which would be imperative for selecting the appropriate MVI model for a given data set. The clustering is performed on median imputed data where the separation between data clusters diminishes with increasing missingness imputed by default median values. Therefore, the performance of our cluster-based MVI approach is limited by the percentage of missingness.

\section {Conclusions}

This paper aims to improve one of the most successful MVI algorithms, MICE, by modifying its linear mapping between observed and missing values using ensemble learning, deep learning, and cluster information. Our extensive analyses across a wide range of missingness, six data sets, and three missingness types conclude that ensemble learning and cluster information together consistently improve the imputation accuracy of the MICE framework. However, the MICE-based models are vulnerable to high missingness and the MNAR type compared to non-MICE models. This shortcoming of MICE is unknown because MVI studies are often limited up to 50\% missingness and MAR/MCAR types. The deep learning-based MICE framework may be superior with specific data sets and missingness types, including MNAR, but at substantially higher computational costs than other models. In the future, the issues listed in the limitation section will be addressed to improve the robustness and explainability of the proposed imputation models.

\section*{Acknowledgements}

Research reported in this publication was supported by the National Library Of Medicine of the National Institutes of Health under Award Number R15LM013569. The content is solely the responsibility of the authors and does not necessarily represent the official views of the National Institutes of Health.

\bibliographystyle{unsrt}
\bibliography{impute}

\end{document}